\documentclass[10pt,twocolumn,letterpaper]{article}

\usepackage{3dv}
\usepackage{times}
\usepackage{epsfig}
\usepackage{graphicx}
\usepackage{amsmath}
\usepackage{amssymb}
\usepackage{color, colortbl}
\usepackage{xspace, xcolor}
\usepackage{booktabs, multirow}
\usepackage[numbers,sort]{natbib}
\DeclareMathOperator*{\argmin}{\arg\!\min}

\definecolor{Gray}{gray}{0.9}

\usepackage{hyperref}

\threedvfinalcopy 

\def\threedvPaperID{218} 

\newcommand{\U}[1]{\underline{#1}}
\newcommand{\B}[1]{\textbf{#1}}
\newcommand{\bl}[2]{\begin{tabular}[l]{@{}c@{}}{#1}\\{#2}\end{tabular}}

\ifthreedvfinal\fi
\begin{document}

\title{Adversarial Self-Supervised Scene Flow Estimation}

\author{Victor Zuanazzi\thanks{Research done while intern at TomTom} \\
University of Amsterdam\\
{\tt\small victorzuanazzi@gmail.com}
\and
Joris van Vugt \\
TomTom, Amsterdam \\
{\tt\small jorisvanvugt@gmail.com}
\and
Olaf Booij \\
TomTom, Amsterdam \\
{\tt\small olaf.booij@tomtom.com}
\and
Pascal Mettes\\
University of Amsterdam\\
{\tt\small P.S.M.Mettes@uva.nl}
}

\maketitle

\begin{abstract}
This work proposes a metric learning approach for self-supervised scene flow estimation. Scene flow estimation is the task of estimating 3D flow vectors for consecutive 3D point clouds. Such flow vectors are fruitful, \eg for recognizing actions, or avoiding collisions. Training a neural network via supervised learning for scene flow is impractical, as this requires manual annotations for each 3D point at each new timestamp for each scene. To that end, we seek for a self-supervised approach, where a network learns a latent metric to distinguish between points translated by flow estimations and the target point cloud. Our adversarial metric learning includes a multi-scale triplet loss on sequences of two-point clouds as well as a cycle consistency loss. Furthermore, we outline a benchmark for self-supervised scene flow estimation: the Scene Flow Sandbox. The benchmark consists of five datasets designed to study individual aspects of flow estimation in progressive order of complexity, from a moving object to real-world scenes. Experimental evaluation on the benchmark shows that our approach obtains state-of-the-art self-supervised scene flow results, outperforming recent neighbor-based approaches. We use our proposed benchmark to expose shortcomings and draw insights on various training setups. We find that our setup captures motion coherence and preserves local geometries. Dealing with occlusions, on the other hand, is still an open challenge.\footnote{This paper was accepted in the 2020 3DV conference.}
\end{abstract}

\section{Introduction}
\label{sec:intro}

This paper introduces an adversarial metric learning approach for estimating 3D scene flow from self-supervision \footnote{https://github.com/VictorZuanazzi/AdversarialSceneFlow.git}. Scene flow estimation constitutes the task of estimating the flow between two unordered 3D point clouds in sequence. The flow information is relevant for a number of applications~\cite{actionflow, flow3dreconstruction, flow3dreconstruction2, flowsegmentation, Behl_2017_ICCV, Vogel_2013_ICCV}. One particular example being self-driving features for the automotive industry.
Several works have addressed this problem from a supervised perspective, where flow vectors are computed from the points of one cloud to the next~\cite{PointFlowNet, FlowNet3D++, HPLFlowNet, FlowNet3D, PointPWCNet}. Since point-level supervision is hard to obtain from real-world data, such approaches are generally trained on synthetic data. Here, we aim for a self-supervised approach to enable a generalization to real-world scene flows.


Recently, Mittal \etal~\cite{SelfSupervisedFlow} and Wu \etal~\cite{PointPWCNet} have introduced self-supervised approaches for scene flow estimation. Common among both approaches is the reliance on flow ground truths in their training setup. Mittal \etal~\cite{SelfSupervisedFlow} advocate a nearest neighbour self-supervision to be used as an intermediary step between supervised training and supervised finetuning. Wu \etal~\cite{PointPWCNet} works under the implicit assumption that a one-to-one mapping between two point clouds exists. That is, the second point cloud is the first point cloud translated by the ground truth flow. We argue that a correspondence assumption between points over time is too strict for real-world applications, where flow vectors are not available and points are measurements re-sampled at each time step. Due to this re-sampling, points can be occluded, and can exit the sensor range.


For real-world scene flow estimation from self-supervision, we should forego the correspondence assumption. Consider for instance a LiDAR sensor on a self-driving vehicle. The sensor generates a new 3D point cloud each time step, with no point-wise correspondence over time. Since the problem is ill-posed, we seek to find a learned approximation of the flow. To that end, we propose to perform self-supervised scene flow estimation on the latent encoding of point clouds. We introduce an adversarial metric learning approach, where a flow extraction network learns to predict a flow field. During training, a separate point cloud embedding network maps the transformed point cloud to a latent space, where we adversarially optimize the flow using a multi-scale triplet loss with cycle consistency. 

We furthermore introduce the Scene Flow Sandbox, a collection of five datasets with increasingly complex scene flows, from individual objects without background to real-world outdoor scenes. On this benchmark, we show the benefit of our approach for self-supervised scene flow estimation, as well as structural open problems. The benchmark and our code is made publicly available to further encourage research in self-supervised scene flow estimation.

\section{Related Work}
\label{sec:relwork}

To tackle the problem of scene flow estimation, a straightforward scheme is to fit a motion model on corresponding primitives on the point cloud. In~\cite{dewan16iros} correspondences are found on triangular meshes which are then fed into a least squares solver. The recently proposed method FLOT~\cite{puy20eccv} goes a step further and uses deep learning and graph matching to find corresponding points directly. Such approaches rely heavily on the assumption of corresponding points in the input point clouds; an assumption we seek to forego.

Building on well-established work in dense optical flow estimation, several recent works have investigated unsupervised and self-supervised image-based scene flow. These works learn flow automatically from sequences~\cite{liu2019selflow} and even tackle occlusions in the learning \cite{Hur_2020_CVPR,meister2017unflow,saxena2019pwoc}. In contrast, our approach is specifically designed for learning scene flow for unordered 3D points in point clouds, rather than grid-structured images. Our triplet loss takes advantage of the permutation invariant nature of point clouds by sampling points, allowing us to automatically learn flow for 3D points.

Building upon the power of deep learning to learn implicit representations, various architectures have been proposed that learn an abstract representation of a 3D point cloud, which can then be used for scene flow estimation amongst others.
The method proposed in~\cite{ushani17icra}
uses a 3D voxel space to represent point clouds, learning a 3D ConvNet to discriminate between differently moving objects. In~\cite{FlowNet3D} FlowNet3D was proposed, with Pointnet++ as the main building block~\cite{PointNet++}.
The follow-up work FlowNet3D++~\cite{FlowNet3D++} shows the benefit of using point-to-plane and cosine distances as auxiliary geometric losses. 
Another architecture, PointPWC-net~\cite{PointPWCNet},
is based on three components: coarse to fine pyramids, warping layers, and cost volumes. PointConv is used for feature extraction~\cite{pointconv}.
The authors of~\cite{HPLFlowNet} propose HPLFlowNet that makes use of 2D permutohedral lattices, as proposed in the SplatNet method~\cite{splatnet}.
Common amongst these approaches is the need for supervised setups for learning scene flow,
where the end-point-error given the ground truth flow is minimized directly via an $\ell_2$ loss. Contrarily, we aim for a self-supervised approach to avoid the need for manual annotations when moving towards real-world scenarios.

Recently, two self-supervised setups have been proposed. 
PointPWC-net~\cite{PointPWCNet} proposes, besides its supervised losses, a self-supervised setup. They make use of the chamfer distance between point clouds as their main training signal. The chamfer distance is combined with a smoothness constraint and Laplacian regularization, which enforces consistent local plane normals. 
More recently, \cite{SelfSupervisedFlow} proposes to endow the FlowNet3D architecture \cite{FlowNet3D} with a nearest neighbour and cycle consistency loss. The flow model is trained in three steps. First, it is trained with supervision on FlyingThings3D~\cite{flyingthings3d}, then trained on NuScenenes and Lyft~\cite{nuscenes2019, lyft2019} using the self-supervised losses, and further finetuned - with supervision - on KITTI Scene Flow~\cite{kitti}. 

Both self-supervised approaches are based on nearest-neighbour based distances to approximate motion, relying again on the assumption of corresponding points in the input point clouds. In this work, we propose an adversarially learned loss that does not have this drawback. On top of that we aim for a fully self-supervised approach, which does not suffer from the domain shift between synthetic and real data and does not require supervised finetuning.



\section{Method}

\begin{figure*}
    \centering
    \includegraphics[width=.99\linewidth]{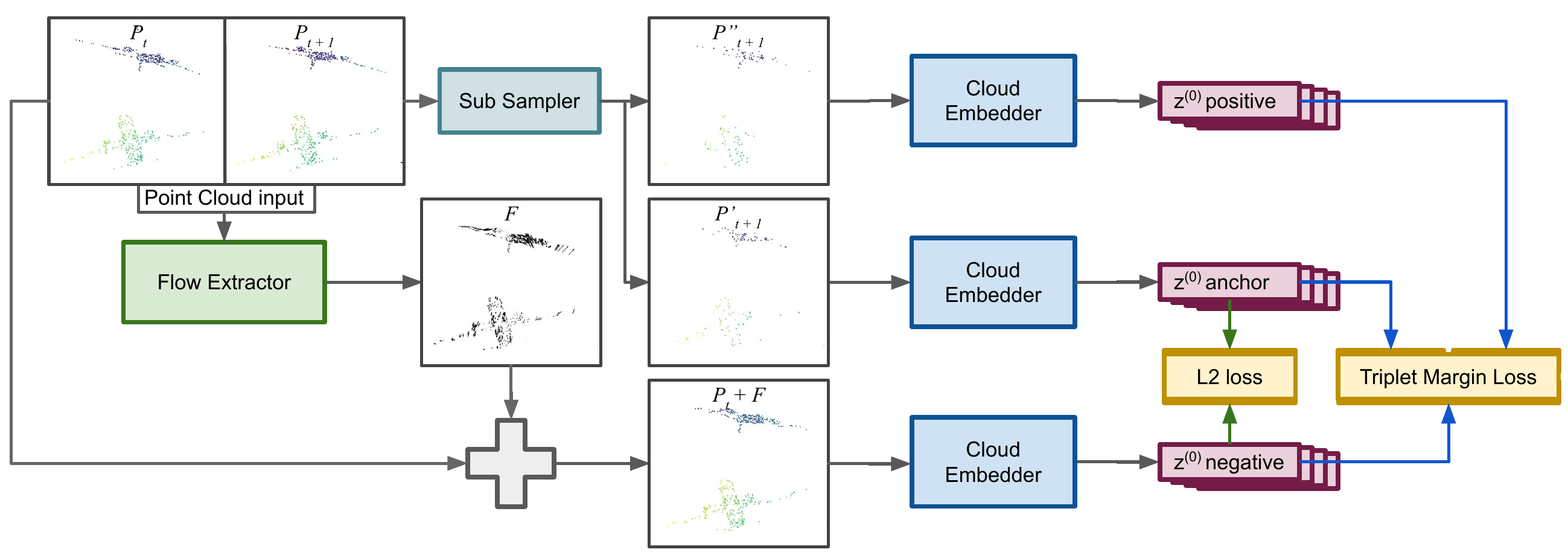}
    \caption{Overview of adversarial metric learning for scene flow estimation. The flow extractor predicts the scene flow from a point cloud at time $t$ and $t+1$. The predicted flow is used to transform the first point cloud into the second. The cloud embedder maps each point cloud to a latent representation, where we employ and adversarial version of the triplet loss to optimize both networks. Blue arrows illustrate the loss used to train embedding function $g_\phi$ and green arrows illustrate the loss used to train flow extractor $h_\theta$.}
    \label{fig:overview}
\end{figure*}
\label{sec:method}
In this section, we will formalize self-supervised scene flow estimation on point clouds and describe our adversarial metric learning approach to the problem. An overview of our method is depicted in Figure~\ref{fig:overview}.

\subsection{Scene Flow}
Formally, the problem of scene flow estimation can be defined as finding the flow vector $\vec{f}$, such that
\begin{equation}
    \vec{p}_t + \vec{f} = \vec{p}_{t+1},
    \label{eq:flow}
\end{equation}
where vector $\vec{p}_t$ represents a point in point cloud $P$ at time $t$. Importantly, $\vec{p}_{t+1}$ is generally never measured in applications such as a car-mounted LiDAR sensor. The second measured point cloud $P_{t+1}$, is constructed from a new sampling of the scene. As a result, $\vec{p}_{t+1}$ can be occluded, outside the field-of-view of the sensor, or simply not exactly sampled again. Despite the ill-posed nature of scene flow estimation, self-supervised learning can be employed to learn scene flow from sequences of point clouds. The general setup is as follows,
\begin{equation}
    \argmin_{\theta} \, d(\tilde{P}_{t+1}, P_{t+1}),
\end{equation}
where $d(\cdot,\cdot)$ is a distance function between two point clouds,
\begin{equation}
    \tilde{P}_{t+1} = P_t + h_\theta(P_t, P_{t+1}),
    \label{eq:ptilde}
\end{equation}
and $h$ is a neural network parameterized by $\theta$ that predicts scene flow. Different from nearest neighbour-based methods for modelling $d(\cdot,\cdot)$ on the point-level, we propose to measure distances on the manifold-level.


\subsection{Adversarial metric learning}
To enable self-supervised scene flow estimation in real-world settings, where point clouds are resampled at each time step, we propose to compute the distance between two point clouds on a latent space as:
\begin{equation}
    d(P, Q) = ||g_\phi(P) - g_\phi(Q)||_2,
\end{equation}
where $g$ is a neural network with parameters $\phi$ which independently maps point clouds $P, Q \in \mathbb{R}^{N \times 3}$ to a vector $\vec{z}$ in latent space. Our proposed adversarial metric learning consists of four components: \emph{(i)} a triplet loss with anchor and positive sampling, \emph{(ii)} a cycle consistency loss, \emph{(iii)} multi-scale triplets for global and local consistency, and \emph{(iv)} adversarial optimization.
\\\\
\textbf{Triplet loss.}
The embedding function $g_\phi$ is optimized using the triplet margin loss \cite{TripletMargin}
\begin{equation}
    \mathcal{L}_\text{tri}(\vec{z}_a, \vec{z}_p, \vec{z}_n) = \max(||\vec{z}_a - \vec{z}_p|| - |||\vec{z}_a - \vec{z}_n|| + m, 0),
\end{equation}
where margin $m$ is generally set to $1$. To obtain suitable triplets from two consecutive point clouds, we propose to first randomly partition point cloud $P_{t+1}$ into two new clouds $P'_{t+1}$ and $P''_{t+1}$. By design, these two clouds represent the same scene and can serve as the anchor $z_a=g_\phi(P'_{t+1})$ and positive $z_p=g_\phi(P''_{t+1})$ for the triplet loss. The negative $z_n=g_\phi(\tilde{P}_{t+1})$ is computed from the cloud at time $t$ transformed by the estimated flow as in Equation (\ref{eq:ptilde}). With this approach, only two frames are needed to enable triplet-based self-supervision on the latent manifold.
\\\\
\textbf{Cycle consistency.}
To further enhance the self-supervision, we add a cycle consistency term to the loss function of the flow extractor. Intuitively, if we reverse the point cloud sequence the scene flow should be opposite to the original, i.e., $h_\theta(P_t, P_{t+1}) = -h_\theta(\tilde{P}_{t+1}, P_t)$. Note that we have to use the estimated next cloud $\tilde{P}_{t+1}$ instead of the measured cloud $P_{t+1}$, as we can not assume any correspondences between clouds from different time steps. Prior work \cite{FlowNet3D, SelfSupervisedFlow} minimizes the $\ell_2$ norm of the difference between the forward $\vec{f}$ and the backward flow $\vec{b}$. A degenerate solution for this formulation is that the network will predict flow vectors with small magnitudes. To solve this problem we take the cosine similarity into account to encourage the network to predict flow in the right direction. Our cycle consistency loss is as follows:
\begin{equation}
    \mathcal{L}_\text{cc} = \sum^N_i || \vec{f}_i + \vec{b}_i ||_2 + \frac{\vec{f}_i \cdot \vec{b}_i}{||\vec{f}_i||_2 \times ||\vec{b}_i||_2},
\end{equation}
where $f_i$ and $b_i$ are respectively the forward and backward flow of the $i$-th point.
\\\\
\textbf{Multi-scale triplets.} To ensure both global and local consistency between the predicted and measured point cloud, we compute the triplet loss at each layer in the cloud embedding network. The intuition is that shallow layers model local structures, whereas deeper layers model higher level properties. The multi-scale triplet loss is defined as a weighted sum of the triplet loss at each layer
\begin{equation}
    \mathcal{L}_\text{mst} = \sum_{l=0}^L \gamma(l) \mathcal{L}_\text{tri}(\vec{z}_a^{(l)}, \vec{z}_p^{(l)}, \vec{z}_n^{(l)}),
\end{equation}
where $l=0$ refers to the last layer, and $\vec{z}^{(l)}$ is the feature vector of layer $l$ obtained via average pooling. We weight the multi-scale triplets by emphasizing triplets deeper in the network higher as:
\begin{equation}
    \gamma(l) = \frac{1}{\sqrt{l + 1}}.
\end{equation}
\\\\
\textbf{Adversarial optimization.}
Using the multi-scale triplets and the cycle consistency, we propose to optimize the self-supervised scene flow estimation task in an adversarial manner. We start by defining the distances between the embeddings of the anchor, and the positive and negative samples at each layer

\begin{figure*}
    \includegraphics[width=0.99\linewidth]{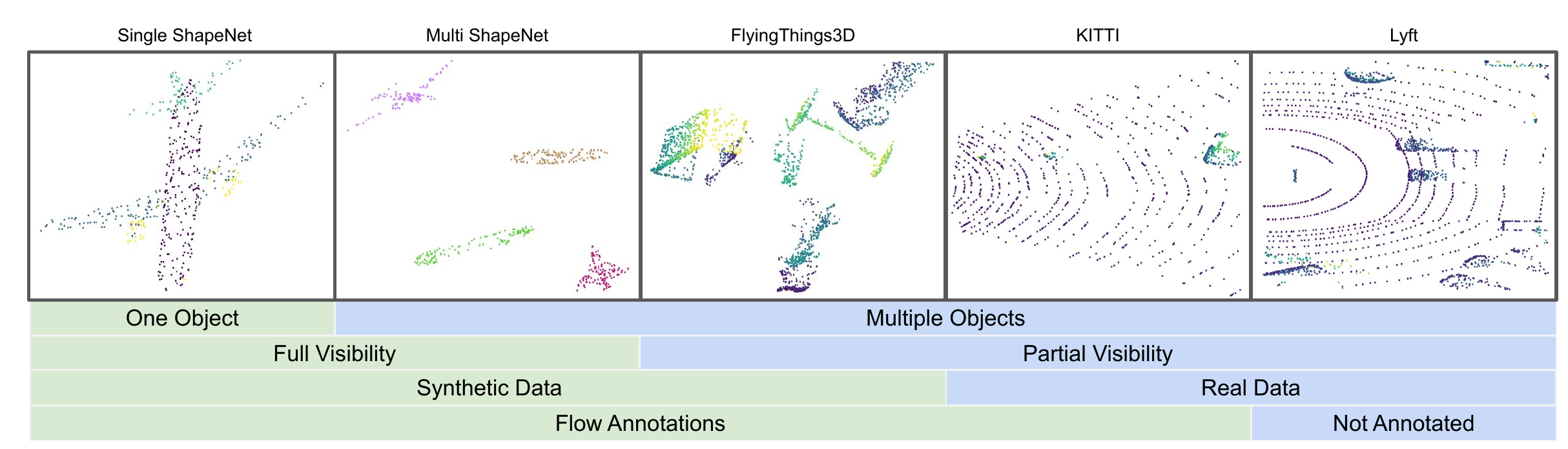}
    \caption{Illustration and summarization of the differences between datasets.}
    \label{img:sandbox}
\end{figure*}

\begin{align}
    r_p^{(l)} &= ||\vec{z}_a^{(l)} - \vec{z}_p^{(l)} ||_2 \\
    r_n^{(l)} &= ||\vec{z}_a^{(l)} - \vec{z}_n^{(l)} ||_2.
\end{align}
Using the distances, the loss for flow extractor $h_\theta$ is given as the difference between the embeddings of the flow-transformed point cloud and the cycle consistency loss:
\begin{equation}
    \mathcal{L}_h = \sum_{l=0}^L \gamma(l) r_n^{(l)} + \lambda\mathcal{L}_{cc},
\end{equation}
with hyperparameter $\lambda$. For the embedding network $g_\phi$ we have the following loss:
\begin{equation}
    \mathcal{L}_g = \sum_{l=0}^L \gamma(l)\max(r_p^{(l)} - r_n^{(l)} + m, 0).
\end{equation}
The two loss functions create an adversarial set-up, since the flow extractor minimizes $r_n^{(l)}$, and the cloud embedder maximizes the same quantity. The networks are optimized in rounds. While the parameters of the flow extractor are kept frozen, we train the cloud embedder for one batch. Then we use the same batch of data to train the flow extractor while keeping fixed the parameters of the cloud embedder. In a Nash equilibrium, the latent representation of the flow-transformed point cloud is expected to be indistinguishable from that of the measured point cloud.

\label{sec:aml}
\section{Experimental setup}

In this section, we will outline the experimental setup, including the Scene Flow Sandbox, the implementation details, and the evaluation metrics.

\subsection{Scene Flow Sandbox}
In order to have an environment to test the potential and limits of self-supervised scene flow estimation, we introduce the Scene Flow Sandbox.
The benchmark consists of five datasets designed to study individual aspects of flow estimation in progressive order of complexity, from a single object in motion to real-world scenes. Failures seen on the synthetic datasets are expected to also surface on real data. We are mostly interested in exploring the aspects summarized in Figure~\ref{img:sandbox}, namely the number of objects, the level of visibility, the use of synthetic or real data, and the level of annotations. 

The first aspect we are interested in studying is related to the number of objects in a scene. The simplest dataset -- \textbf{Single ShapeNet} -- is made of one moving, fully visible, object. The geometry of the object does not suffer major changes between two frames. \textbf{Multi ShapeNet} increases the complexity of the scene by adding multiple objects to the scene. Even though the geometry of each object must still be kept consistent, the geometry of the scene may drastically change. In both datasets, a new set of points is sampled at each time step, the point clouds are taken from ShapeNet \cite{shapenet} and then modified by a transformation matrix.

The second aspect regards how the visibility of a scene impacts its complexity. In a fully visible scene, the model is aware of all the parts of the objects at all times. \textbf{FlyingThings3D} \cite{flyingthings3d} introduces partial observability, which means objects do self-occlude, they may occlude each other, and parts of objects may leave and enter the field of view. We are interested in studying how, and if, a model learns useful motion priors for occluded regions. 
The third aspect is the inherent differences between synthetic and real data collected by LiDAR. KITTI Scene Flow \cite{kitti} has 200 annotated scenes, far too few for the training of sophisticated models. We use \textbf{Lyft} \cite{lyft2019} as the source of unsupervised training data and \textbf{KITTI} for evaluation and testing. Both datasets use points that are in front of the car and inside a cube with sides of 30 meters.

From the least to the most complex dataset, the sandbox was tailored to facilitate insightful experimentation. 

\subsection{Implementation details}

For our experiments we use FlowNet3d \cite{FlowNet3D} and PointPWC-net \cite{PointPWCNet} as flow extractors. The cloud embedder uses three PointNet++ Set Abstraction Layers \cite{PointNet++} with mean pooling and three blocks of fully connected layers. The final feature vector has 4096 dimensions.
We use the Adam optimizer with a learning rate of $5 \times 10^{-4}$ for the flow extractor and $5 \times 10^{-5}$ for the cloud embedder. For both models, the momentum term is set to zero, and the weight decay set to $10^{-4}$. Both learning rates are multiplied by $0.75$ after 20 epochs of no EPE improvement. 

\subsection{Metrics}

Flow estimation is a regression task. In alignment with previous work \cite{FlowNet3D, FlowNet3D++, PointPWCNet, HPLFlowNet, SelfSupervisedFlow}. The quality of the estimations is measured using the following metrics. 

\emph{End Point Error} (EPE): the average Euclidean distance between the target and the estimated flow vectors. It is the main metric we are interested in improving. 

\emph{Accuracies}: the percentage of flow predictions that are below a threshold. The threshold has two criteria, meeting one of them is sufficient. \emph{Acc 01}: the prediction error is smaller than $0.1$ meter or $10\%$ of the norm of the target. \emph{Acc 005}: the prediction error is smaller than $0.05 $ meter or $5\%$ of the norm of the target.



\section{Results}
\label{sec:exps}

\begin{table}
    \begin{center}
        \begin{tabular}{cccccc}
        \toprule
        \multicolumn{3}{c}{\textbf{Cycle losses}} & \textbf{EPE} & \textbf{Acc 01} & \textbf{Acc 005}\\
        Cosine & MSE & $\ell_2$ & & &\\
        \midrule
        & & & 0.4920 & 2.85\%  & 0.39\%  \\
        $\checkmark$ & & & 0.4302 & 5.44\%  & 0.83\%  \\
        & $\checkmark$ &  & 0.4405 & 4.37\%  & 0.63\%  \\
        & & $\checkmark$ & 0.3786 & \B{12.67}\%  & \B{3.02}\%  \\
        $\checkmark$ & $\checkmark$ & & 0.4200 & 6.56\%  & 1.08\%  \\
        \rowcolor{Gray}
        $\checkmark$ & & $\checkmark$ & \B{0.3497} & 10.27\% & 1.78\% \\
        \bottomrule
        \end{tabular}
    \end{center}
    \caption{Ablation study I: cycle consistency. We evaluate the cycle consistency in our approach on Multi ShapeNet. All variants improve the self-supervised scene flow estimation, with a combination of cosine and $\ell_2$ preferred as it achieves the lowest EPE.}
    \label{tab:ablation_cycleloss}
\end{table}

In this section, we report our experiments, results, and insights. We start with two ablation studies, investigating the effect of cycle consistency and the multi-scale triplet loss in our adversarial metric learning. Then, we evaluate the effect of removing the correspondence assumption, prevalent in the current self-supervised approaches, during training and testing. Afterwards, we perform a comparative evaluation to the state-of-the-art in self-supervised scene flow estimation and we finalize the experiments with an evaluation on self-supervised scene flow estimation on real data.


\subsection{Ablation I: Cycle Consistency Loss}

\begin{table}
    \begin{center}
        \begin{tabular}{cccc}
            \toprule \textbf{Multi-scale factors} & \textbf{EPE} & \textbf{Acc 01} & \textbf{Acc 005}\\
            \midrule
            None  & 0.4043 & 5.60\%  & 0.76\%  \\
            \rowcolor{Gray}
            $\frac{1}{\sqrt{l + 1}}$   & \B{0.3497} & \B{10.27}\% & \B{1.78}\%  \\
            $\frac{1}{l + 1}$          & 0.3850 & 7.57\%  & 1.12\%  \\
            $\frac{1}{(l + 1)^2}$      & 0.4137 & 7.33\%  & 1.23\%  \\ 
            \bottomrule
        \end{tabular}
    \end{center}
    \caption{Ablation study II: multi-scale triplet loss. We evaluate on Multi ShapeNet using FlowNet3D. The scaling factors are functions that use the level $l$ of the activations. The activations are organized from last to first. Baseline (none) only uses the last activation. Performing triplet losses at multiple scales directly improves the results over using the last activations only.}
    \label{tab:mstloss}
\end{table}

For the first experiment, we evaluate the contribution of the cycle consistency loss in our approach.
We consider five cycle consistency variants, namely cosine similarity, MSE, and $\ell_2$.
Table~\ref{tab:ablation_cycleloss} shows the results of the ablation study on Multi Shapenet. The best EPE results are achieved using a combination of the cosine similarity and $\ell_2$. All variants improve over the baseline without cycle consistency. Unstable optimization is a common problem in adversarial training. We find that enforcing cycle consistency played an important role in keeping the training stable.

\subsection{Ablation II: Multi-scale triplet loss}
As part of our approach, we perform advservarial learning on the latent manifold at multiple scales. Here, we evaluate its effect on Multi Shapenet
We use the multi-scale triplet loss in or training setup. We compare the use of the last feature vector (scaling is set to zero) against three methods for scaling the intermediary feature vectors.  For all three methods, the scaling of the last feature vector is $1$ and the shallowest layer is the most down-scaled. 
Table~\ref{tab:mstloss} shows the results. The use of the correct scaling method brings at least a $10\%$ improvement when compared to the second-best. The multi-scale triplet loss is the main loss used in our setup. By using multiple feature vectors we allow our latent space to encode both local and global geometric properties of the point cloud. 

\subsection{Correspondence and re-sampling}
\label{sec:mechanism}

\begin{table}
    \begin{center}
        \begin{tabular}{lcccc}
            \toprule
            & \multicolumn{4}{c}{\textbf{Train mechanism $\rightarrow$ Test mechanism}}\\
            & C $\rightarrow$ C & C $\rightarrow$ R & R $\rightarrow$ C & R $\rightarrow$ R\\
            \midrule
            HPLFlow     & 0.0948 & 0.3997 & 0.1965 & 0.2598 \\
            PointPWC-net   & 0.0575 & 0.4747 & 0.1701 & 0.2644 \\
            PointPWC-net$^{\star}$  & 0.0965 & 0.4888 & 0.3455 & 0.5502 \\
            \bottomrule
        \end{tabular}
    \end{center}
    \caption{Experiment on FlyingThings3D showing the effect on EPE using the correspondence (C) and re-sampling (R) mechanisms during training and testing. The $\star$ denotes self-supervised evaluation. We find that relying on correspondences during training results in poor results when such assumptions are no longer met during evaluation.}
    \label{tab:samplevscorrespondence}
\end{table}

By relying on learning on the manifold, we avoid relying on explicit correspondences between individual points to estimate the scene flow. Here, we quantify the effect of relying on explicit correspondences. We make the distinction between the correspondence mechanism, where each point has a unique point at the next timestamp, and the re-sampling mechanism, where a new point cloud is obtained at each new time stamp. We evaluate all for combinations of either mechanism used during training and/or testing. We have used the publicly available code-base \cite{hplflow_repo, pointpwc_repo} and reported set of hyper-parameters to reproduce the work of \cite{HPLFlowNet, PointPWCNet}. The models were trained on FlyingThings3D. All models were evaluated on both the correspondence and re-sampling mechanisms.

Table~\ref{tab:samplevscorrespondence} shows the impact the different mechanisms make on the scene flow performance.
The correspondence mechanism simplifies the task of scene flow estimation. From Table~\ref{tab:samplevscorrespondence} we observe that the re-sampling training generalizes reasonably to the correspondence evaluation, yet, the opposite does not happen. This finding is not surprising. The correspondence mechanism artificially carries over occlusions independently of the motion of the objects. The re-sampling mechanism, on the other hand, allow for entire objects to disappear in-between frames. This limits the information available for flow information. For self-supervised learning, we also find that PointPWC-net is not well suited for training on data that is re-sampled each timestamp. We recommend that re-sampling is used explicitly in scene flow estimation, whether it be supervised or self-supervised, as this matches real-world applications best.




\subsection{Scene flow estimation comparison}
\label{sec:exp_synthetic}
In the fourth experiment, we compare our method with two current approaches on the synthetic datasets of the Scene Flow Sandbox.  We train FlowNet3D and PointPWC-net using our adversarial metric learning. The self-supervised approaches include the FlowNet3D trained from scratch with the self-supervised losses proposed by Mittal \etal \cite{SelfSupervisedFlow} and PointPWC-net trained with the self-supervised losses proposed by Wu \etal \cite{PointPWCNet}. We also include a supervised comparison to FlowNet3D \cite{FlowNet3D} to see how close our self-supervised results are to the supervised upper bound.

\begin{table}
    \centering
    \begin{tabular}{lccc}
         \toprule
         & \bl{Single}{ShapeNet} & \bl{Multi}{ShapeNet} & \bl{Flying}{Things3D} \\
         \midrule
         \rowcolor{Gray}
         \textbf{Supervised} & & &\\
         Liu \etal \cite{FlowNet3D}               & 0.0567 & 0.1524 & 0.1838 \\
         \midrule
         \rowcolor{Gray}
         \multicolumn{4}{l}{\textbf{Self-supervised} (FlowNet3D)}\\
         Mittal \etal \cite{SelfSupervisedFlow} & 0.3911     & 0.8242     & 0.7538\\
         \textbf{This paper} & \B{0.1287} & \B{0.3497}     & \B{0.5629}\\
         \midrule
         \rowcolor{Gray}
         \multicolumn{4}{l}{\textbf{Self-supervised} (PointPWC-net)}\\
         Wu \etal \cite{PointPWCNet} & 0.3895     & 0.8439     & 0.7540 \\
         \textbf{This paper} & \B{0.1824}     & \B{0.2920} & \B{0.5270}\\
         \bottomrule
    \end{tabular}
    \caption{Comparison of different methods for scene flow estimation on the synthetic datasets of the Scene Flow Sandbox. Only EPE is reported. The best metrics of self-supervised methods are reported in bold. The underlined metrics indicate the best overall performance, regardless of the training method.}
    \label{tab:synthetics}
\end{table}

Table~\ref{tab:synthetics} summarizes the results. We find that both existing self-supervised approaches perform similar on the three dataset of our Scene Flow Sandbox. Regardless of the used flow extractor, we outperform the baselines. On Multi ShapeNet, we obtain an EPE of 0.3497 compared to 0.8242 for \cite{SelfSupervisedFlow} when using FlowNet3D. When using PointPWC-net, we obtain an EPE of 0.2920, compared to 0.8439 for \cite{PointPWCNet}. The relative difference with the baselines decreases as the datasets become more challenging, indicating that self-supervised scene flow estimation is anything but a solved task.





\begin{figure*}
    \centering
    \frame{\includegraphics[width = .49\linewidth]{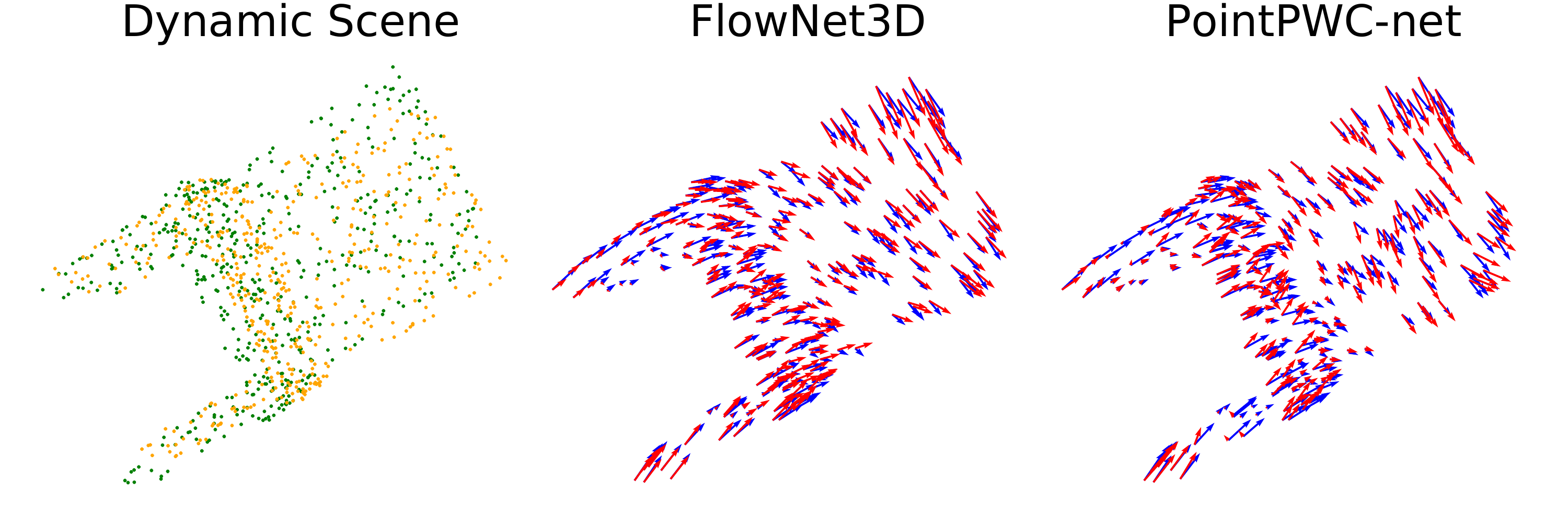}}
    \frame{\includegraphics[width = .49\linewidth]{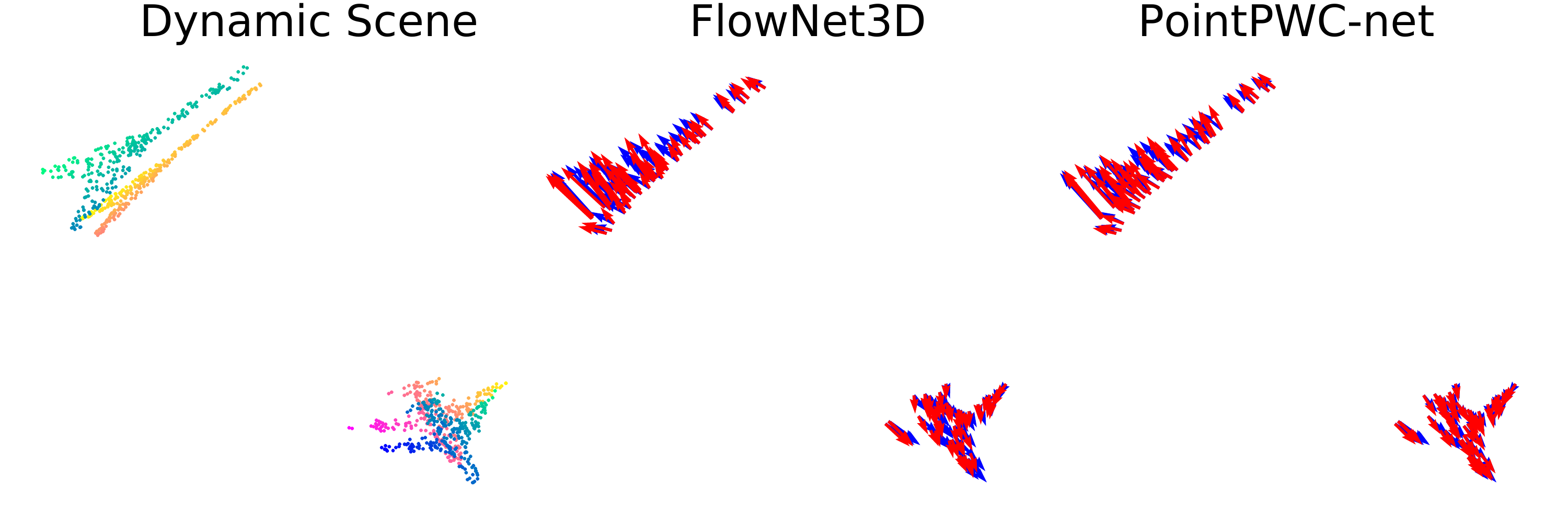}}
    \frame{\includegraphics[width = .99\linewidth]{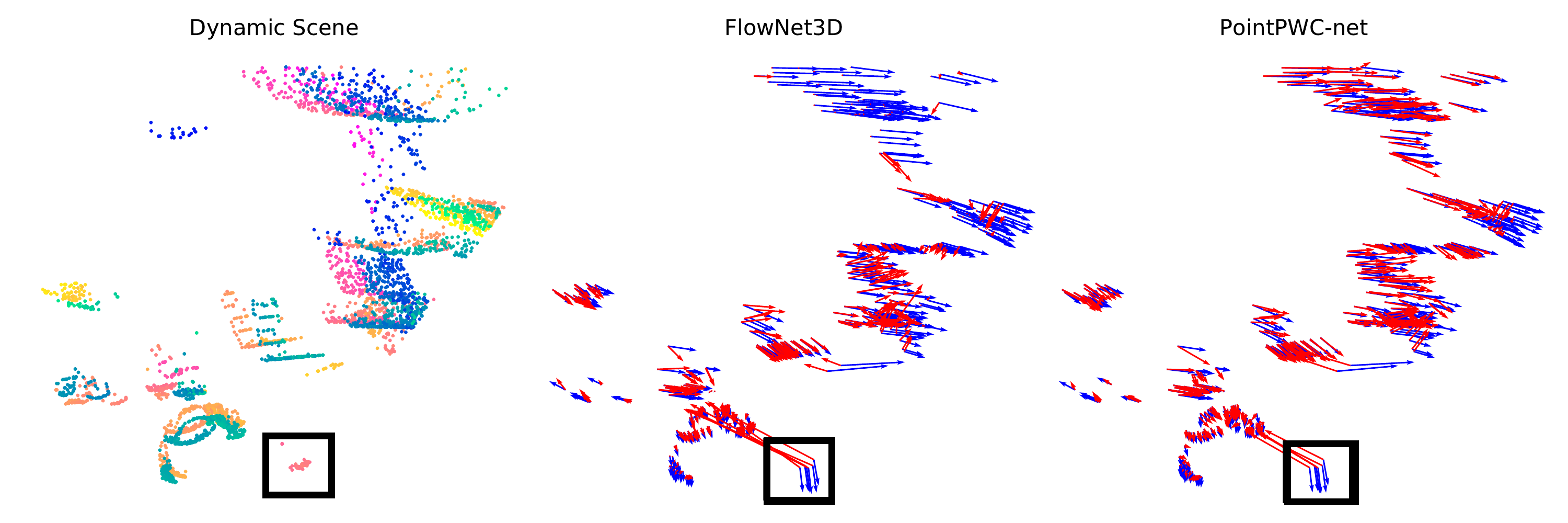}}
    \caption{Qualitative results from adversarial metric learning on Single ShapeNet (top left), Multi ShapeNet (top right) and FlyingThings3D (bottom). Points colored with cold colors belong to frame 1 and warm colors to frame 2. The ground truth vectors are shown in blue and the predicted vectors in red. Black square shows an object that leaves the field of view. Whereas the flow estimations on partially visible scenes are still to be improved, the models trained with out setup learn motion from only two consecutive point clouds.}
    \label{img:syntetics}
\end{figure*}

To understand what keeps our proposed training setup from bridging the gap with the supervised setup, we perform a qualitative analysis in Figure~\ref{img:syntetics}. On Single ShapeNet and Multi ShapeNet, both models perform similarly well. The geometries of the objects are kept and the independent motion of different objects in one scene is well captured. However, the occlusions incorporated by FlyingThings3D seem rather problematic for our training setup. FlowNet3D and PointPWC-net display one common failure in their flow predictions. The black square highlights an object leaving the field of view. Instead of guessing a would-be location for the object, the flow extractors merge it with the nearest object in the scene. These qualitative results indicate the open challenges that remain for self-supervised scene flow estimation.



\subsection{Self-supervised scene flow on real data}

The primary motivation of this work is to perform scene flow estimation on real data. In the fifth experiment, we compare our approach to the self-supervised methods proposed by \cite{SelfSupervisedFlow, PointPWCNet} when trained directly with real data. We also compare to a supervised baseline \cite{FlowNet3D}. 

The supervised baseline is trained on FlyingThings3D and finetuned on KITTI with and without ground, as proposed by \cite{FlowNet3D}. The works of \cite{SelfSupervisedFlow, PointPWCNet} rely on the ground truth flow in different ways. We only show results that do not make use of target flow vectors at training time. The work of \cite{SelfSupervisedFlow} proposes a three-step training. Supervised training a FlowNet3D using FlyingThings3D, followed by self-supervised training on a non-annotated dataset and finally fine-tuning the model on KITTI Scene Flow \cite{kitti}. We trained FlowNet3D from scratch on Lyft, using the loss functions proposed by \cite{SelfSupervisedFlow}, and did not perform the finetuning step. The work of \cite{PointPWCNet} assumes the correspondence mechanism. We trained the PointPWC-net on Lyft, which inherently does not have the correspondence mechanism. All methods are evaluated on the test split of KITTI, with and without ground plane. 

\begin{table*}
    \centering
    \begin{tabular}{cllccc}
        \toprule
        \textbf{Dataset} & \textbf{Training method} & \textbf{Flow extractor} & \textbf{EPE} & \textbf{Acc 01}  & \textbf{Acc 005} \\ 
        \midrule
                         & Supervised \cite{FlowNet3D} *     & FlowNet3D         &  \U{0.1729} & \U{57.68}\% & \U{22.73}\% \\
                         & Self-Supervised \cite{SelfSupervisedFlow} & FlowNet3D & 1.0903 & \B{9.81}\%  & 3.08\% \\ 
        KITTI            & Self-Supervised \cite{PointPWCNet} & PointPWC-net     & 1.1493 & 8.28\%  & \B{6.86}\% \\
                         & Adversarial Metric & FlowNet3D                        & \B{0.9673} & 3.01\%  & 0.76\% \\
                         & Adversarial Metric & PointPWC-net                     & 1.0497 & 3.41\%  & 1.02\% \\ \hline
                         & Supervised \cite{FlowNet3D} *      & FlowNet3D        &  \U{0.1880} & \U{52.12}\% & \U{22.81}\% \\
                         & Self-Supervised \cite{SelfSupervisedFlow} & FlowNet3D & 0.7002 & 5.05\%  &  1.43\% \\ 
        KITTI No Ground  & Self-Supervised \cite{PointPWCNet} & PointPWC-net     & 0.9472 & \B{6.07}\%  & \B{3.74}\% \\
                         & Adversarial Metric & FlowNet3D                        & 0.6733 & 5.82\%  & 1.03\% \\
                         & Adversarial Metric & PointPWC-net                     & \B{0.5542} & 5.58\%  & 1.45\% \\
                         \bottomrule
    \end{tabular}
    \caption{Comparison of different methods for scene flow estimation on KITTI. The best metrics of self-supervised methods are reported in bold. The underlined metrics indicate the best overall performance, regardless of the training method. * Trained on FlyingThings3D and finetuned on KITTI.}
    \label{tab:reals}
\end{table*}

\begin{figure*}
    \includegraphics[width=1.05\linewidth]{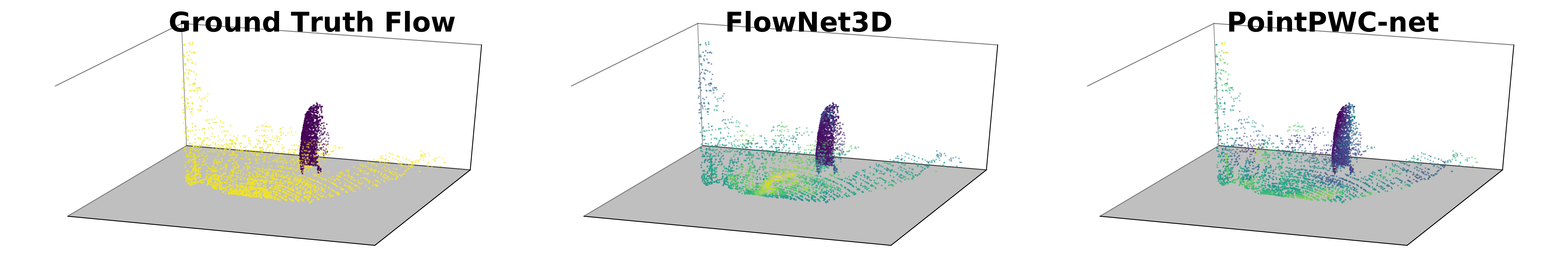}
    \includegraphics[width = 1.05\linewidth]{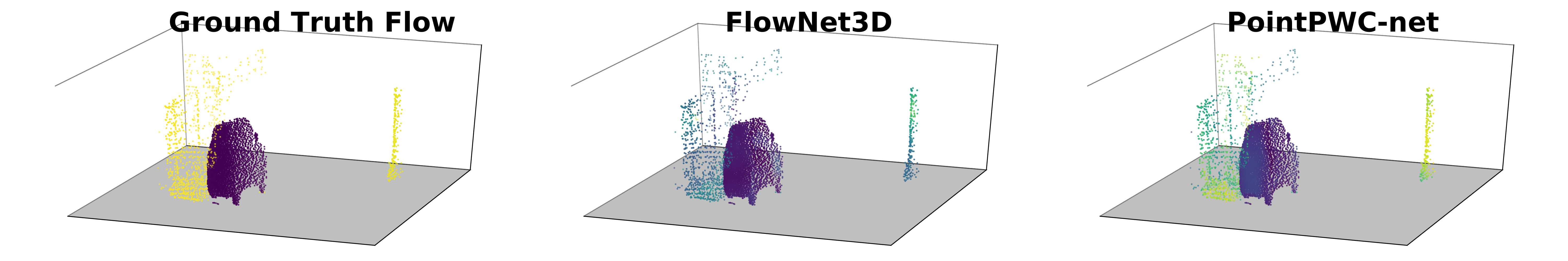}
    \caption{Results on KITTI with ground (top) and without ground (bottom) using FlowNet3D and PointPWC-net. Flow vectors are indicated in the coloring. The ground truth flow is shown on the leftmost column, the predictions from FlowNet3D are shown in the center and the predictions from PointPWC-net are shown in the rightmost column. Our self-supervised method approximates the motion of the vehicle relatively well while deforming the static objects.}
    \label{img:kitti}
\end{figure*}

Table~\ref{tab:reals} summarizes the results. The supervised training followed by finetuning is still state-of-the-art. Generally, our method performs best among the self-supervised baselines in terms of EPE. The gap between supervised and self-supervised, however, is considerably larger than for the synthetic datasets. 
Interestingly, the nearest-neighbour based losses achieve higher accuracy than our method, even though their EPE is considerably worse. It means that, on average, the flow estimations of our method are better than of the baselines. However, the best flow estimations of the baselines are better than the best flow estimations from our method. A reason for this is that in our approach, the cloud embedder performs lossy compression from point cloud to latent vectors. The latent vectors may encode relevant global flow information but may ignore fine motion. The nearest neighbour based losses on the other hand, look for local point assignments. On average, the point assignments are a poor proxy of motion. However, a small number of assignments are expected to capture fine local motion.

In general, the self-supervised models perform better when the ground is removed in the evaluation. The ground is a large plane object that gives little information about motion. Our adversarial metric learning approach is the least impacted by the presence of the ground on real data. We attribute that to the metric learned by the cloud embedder. As opposed to the nearest neighbour based distances, the cloud embedder may give different importance to points belonging to different objects. We conjecture it may learn to regard points belonging to the floor as less informative than points belonging to moving objects

Figure~\ref{img:kitti} shows a qualitative example of KITTI with and without ground. In both scenes, there is one moving car and the rest of the scene is made of static objects, such as rail guards walls, and the ground. Both models approximated the motion of the car with relative success. However, neither model estimated well the lack of motion of the static objects. On the left, we see the static objects are in yellow (zero motion), but the estimations are rather colorful. Which indicates locally inconsistent flow estimations. We conjecture the inconsistencies are a consequence of the occlusions created by the movement of the LiDAR sensor as well as of the car in the scene. 





Overall, adversarial metric learning can be used to train different flow extractors on performing scene flow estimation. The flow extractors, however, are not expected to perform well in partially observable scenes. The result makes sense in hindsight. The flow extractor is encouraged to approximate the target point cloud as much as possible. However, it has no means to perform occlusions. Thus it maps the points to a nearby object to approximate the target point cloud, yielding poor flow estimations. The cloud embedder, on the other hand, can easily distinguish between the target and the predicted point cloud by simply spotting missing objects. Further improvements require handling partially observable scenes.

\section{Conclusion}
\label{sec:conclusion}




This paper addressed the problem of scene flow estimation. Where a supervised setup is infeasible as this requires manual annotations for each point at each timestamp for each scene, we advocate a self-supervised approach. We propose an adversarial metric learning approach, which incorporates a multi-scale loss and cycle consistency. We also introduce the Scene Flow Sandbox. The sandbox provides a constructive setup to uncover the potential and structural limitations of self-supervised scene flow estimation. Empirically, we find that our approach obtains state-of-the-art error rates compared to current neighbour-based approaches. While the flow for individual objects can be learned in a self-supervised manner, a lot of work is needed to deal with real-world data, especially in the presence of occlusions. We hope that our approach and sandbox encourages further research into this problem.


{\small
\bibliographystyle{ieee}
\bibliography{3dv.bib}
}

\end{document}